\documentclass[runningheads]{llncs}

\usepackage{eccv}

\usepackage{eccvabbrv}

\usepackage{graphicx}
\usepackage{booktabs}

\usepackage[accsupp]{axessibility}  

\usepackage{hyperref}

\usepackage{orcidlink}

\begin{document}

\title{Visual-Prompt Guided Wildlife Instance-Level Recognition}


\author{Mufhumudzi Muthivhi\inst{1}\orcidlink{0000-0003-0509-6235} \and
Jiahao Huo\inst{1}\orcidlink{0000-0001-6686-2576} \and
Terence van Zyl\inst{1}\orcidlink{0000-0003-4281-630X} \and
Fredrik Gustafsson\inst{2}\orcidlink{0000-0003-3270-171X}}

\authorrunning{Muthivhi et al.}

\institute{University of Johannesburg, South Africa \email{\{mmuthivhi,216045414,tvanzyl\}@uj.ac.za} \and
Linköping University, Sweden \email{fredrik.gustafsson@liu.se}\\}


\maketitle

\begin{abstract}

Fine-grained wildlife re-identification remains a challenging area in research.
Current state-of-the-art approaches apply a detection and re-identification pipeline. 
We propose a one-stage end-to-end detection and re-identification model that performs identity searching within the latent space. 
We adopt DINOv2 for robust spatial geometry and MegaDescriptor for wildlife re-identification. We enhance latent queries with prompt re-identification features. A detection decoder queries the scene latent space to establish object boundaries around the target identity.
Preliminary findings reflect a competitive mean average precision score of 30.584\% compared to the state-of-the-art two stage approach of 44.89\%. Qualitative results depict effective bounding and identification of animal identities.

\end{abstract}

\section{Introduction}
\label{sec:intro}

Generic object detection models categorize wildlife at the species level. Conversely, instance-level wildlife re-identification requires the discrimination of fine-grained visual patterns. The model is tasked with learning from the unique stripe configurations of a zebra or the spot patterns of a cheetah.
Current state-of-the-art approaches deploy an isolated detection and re-identification pipeline using two separate models~\cite{wu2026overcoming}. An object detection model isolates individuals in a scene, crops the bounding box pixels, resizes them to a fixed resolution, and passes them to an independent re-identification feature extractor for similarity matching against a database~\cite{vcermak2024wildlifedatasets}.
Recent advancements in promptable vision foundation models and visual grounding present a compelling alternative~\cite{liu2023grounding}. We could instead view the database as a prompt to query the scene directly to locate a target. However, adapting single-stage visual grounding architectures for instance-level re-identification introduces a training mismatch. A single transformer decoder layer is forced to simultaneously learn the global scenery to extract geometric boundaries of an object and the local texture patterns of a specific identity.
To resolve this, we propose an end-to-end visual prompting task. A frontier foundation model extracts robust spatial features and a specialist re-identification model extracts fine-grained features.
Preliminary findings depict a competitive mean average precision score (mAP) over a two-stage approach. Qualitative results also depict tight bounding boxes around target identities in dense herd, occluded environments and distant subjects. 

\begin{figure}
  \centering
  \centerline{\includegraphics[width=0.8\textwidth]{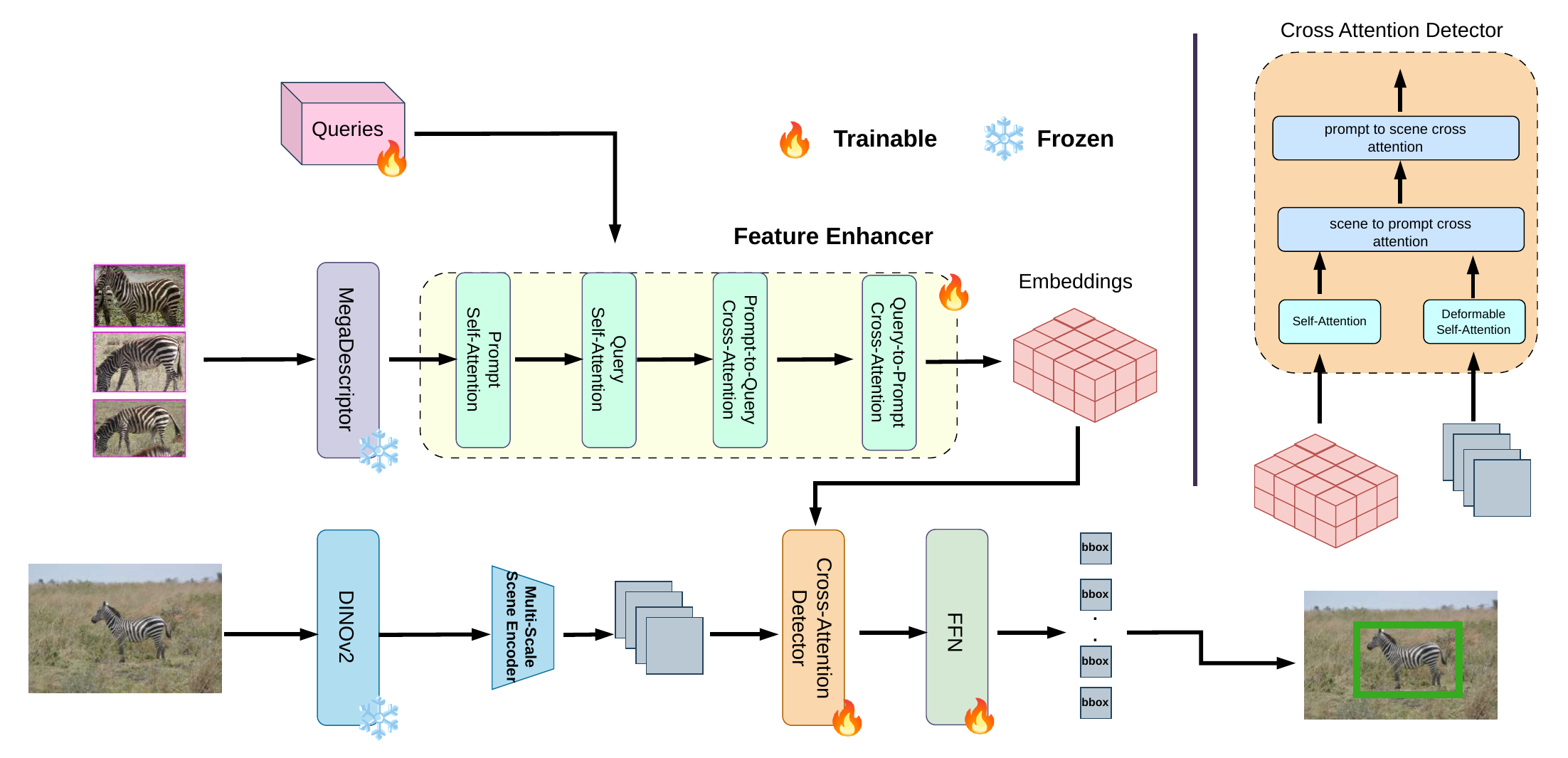}}
   \caption{Overview of the proposed framework. DINOv2 encodes the full scene image to capture spatial features and MegaDescriptor encodes reference crops into identity features used as visual prompts for target matching.}
   \label{fig:framework}
\end{figure}

\section{Methodology}
\label{sec:methodology}

The proposed framework in Figure~\ref{fig:framework} consists of three primary components. First, we extract scene and re-identification features from a frozen encoder. Next, we enhance the object queries with prompt re-identification features to enrich our region proposals. Lastly, we use a cross-attention detector to search for the target identity.

\paragraph{Multi-Scale Scene Encoder}
We use DINOv2~\cite{oquab2023dinov2} to process the full scene image.  To handle extreme scale variations in imagery, we adopt a similar strategy used by Grounding DINO~\cite{liu2023grounding}. We extract intermediate feature maps from layers 3, 6 and 10 within the transformer. These features are projected into a feature pyramid using varying convolutional strides, spatially interpolated to a uniform resolution, concatenated, and subsequently compressed via a $1 \times 1$ convolution. 
\paragraph{Visual prompt}
The visual prompt is obtained through a pretrained fine-grained wildlife re-identification model, MegaDescriptor~\cite{vcermak2024wildlifedatasets}. 

\paragraph{Feature Enhancer}
We initialize a set of learnable object queries as the region proposals. The feature enhancement layer ensures that the queries cross-attend directly with the visual prompt features. We use both self and cross attention to enrich our region proposals (queries) with identity-related information from our visual prompts.

\paragraph{Cross-Attention Decoder}
The enriched region proposals attend to the multi-scale scene features via Deformable Attention~\cite{xia2022vision}. The decoder outputs a set of $N$ cross-attended queries. The cross-attended queries are passed through a feed-forward network to encapsulate the detected animal with a bounding box. An additional objectiveness score is also retrieved, indicating the presence of a valid foreground animal.  

\paragraph{Loss Function}
This work proposes the optimization of four objectives, such that:
\begin{equation}
    \mathcal{L} = \lambda_{obj}\mathcal{L}_{\mathrm{FocalLoss}} + \lambda_{L1} \mathcal{L}_{L1} + \lambda_{GIoU} \mathcal{L}_{GIoU} + \lambda_{\mathrm{div}} \mathcal{L}_{\mathrm{div}}
\end{equation}
is the detection loss. $\mathcal{L}_{FL}$ is the Sigmoid Focal Loss over the objectness score obtained from the $N$ queries. $\mathcal{L}_{L1}$ and $\mathcal{L}_{GIoU}$ is the L1 loss and Generalized Intersection over Union of the predicted and ground truth bounding boxes. $\mathcal{L}_{div}$ encourages diversity of the embeddings of the queries produced by the feature enhancement layer.

\begin{figure}
  \centering
  \centerline{\includegraphics[width=0.9\textwidth]{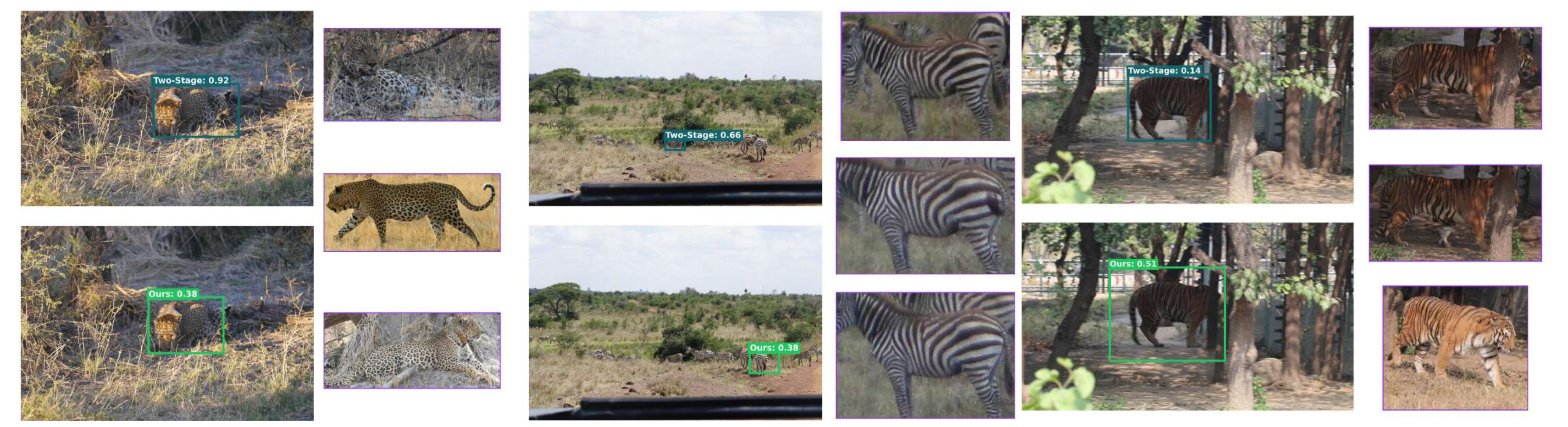}}
   \caption{Qualitative visual grounding results of the proposed architecture. For each scenario, the target identity prompts are displayed on the left, and the scene interrogation results are on the right.}
   \label{fig:Results}
\end{figure}

\section{Experimental Setup}
\label{sec:setup}
\paragraph{Datasets}
We use a curated re-identification dataset from \textit{WildlifeDatasets}~\cite{vcermak2024wildlifedatasets}. We specifically utilize the \textit{GiraffeZebraID}~\cite{parham2017animal} and \textit{ATRW}~\cite{li2019atrw} datasets because they come with the full-scene image and the bounding box metadata. 
\paragraph{Implmentation}
We employ a ViT-S/14 variant of DINOv2 for the multi-scale scene encoder and a MegaDescriptor-T variant for the prompt encoder. Scene images are resized to $518 \times 518$ to align with the DINOv2 patch configuration, while prompt crops are resized to $224 \times 224$. We selectively unfreeze the final transformer block and normalization layers of both encoders during training, allowing their distinct latent manifolds to synchronize. We use PyTorch Lightning for efficient training~\cite{falcon2019pytorch}.
\paragraph{Hyperparameters:}
The network is optimized end-to-end using the AdamW optimizer with a base learning rate of $1 \times 10^{-4}$ and a weight decay of $1 \times 10^{-4}$. We employ a MultiStepLR learning rate scheduler, which decays the rate by a factor of $\gamma = 0.1$. The objectness, identity matching, bounding box L1 distance, and Generalized IoU is set to 2.0, 2.0, 5.0, and 2.0, respectively.

\section{Preliminary Results}
\label{sec:results}

\begin{figure}
  \centering
  \centerline{\includegraphics[width=0.85\textwidth]{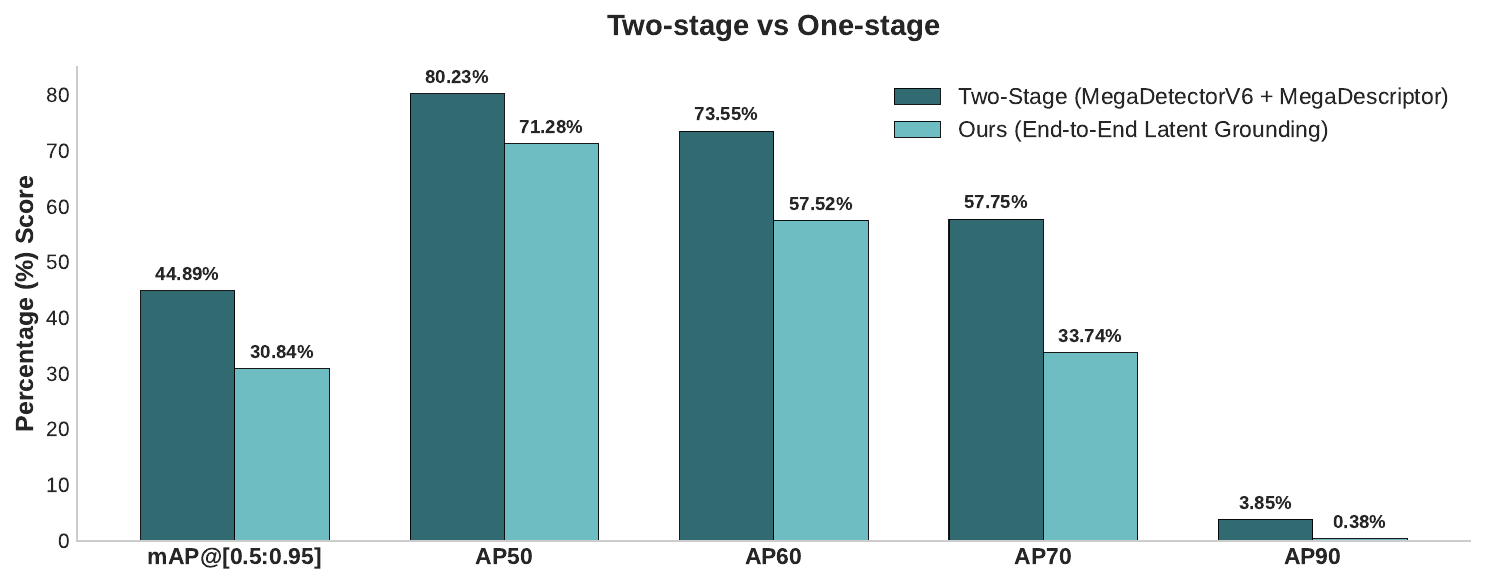}}
   \caption{Our End-to-End Latent Grounding method (one stage) compared to two stage method using MegadetectorV6~\cite{beery2019efficient} with MegaDescriptor~\cite{vcermak2024wildlifedatasets}. The results shown is evaluated on the combined test set from GiraffeZebraID~\cite{parham2017animal} and ATRW~\cite{li2019atrw}}
   \label{fig:Comparison}
\end{figure}

We provide qualitative preliminary findings in Figure~\ref{fig:Results}. The proposed framework produces bounding boxes that encapsulate the target animal. Our model can handle herds, domain shifts, occlusion, distance, and camouflage. 
We present results on retrieval performance (ReID) in Figure~\ref{fig:Comparison} using our end-to-end latent grounding method and compared it to the two-stage approach that uses MegadetectorV6 for detection, crop and followed by MegaDescriptor to embed the cropped images to perform retrieval. The results show the potential of the proposed approach even though it has not yet achieved state-of-the-art performance. However, the proposed approach is much more efficient.

\section{Conclusion}
\label{sec:conclusion}
This paper proposes a framework that combines visual prompt along with global scene features to perform end-to-end object detection together with re-identification. 
The preliminary findings suggest that visual identity prompts can reduce the reliance on traditional two-stage detect and crop followed by ReID pipelines by allowing the model to localize and match a target directly within the full scene image. We believe continued work in this direction could lead to state-of-the-art performance.


%
%
\bibliographystyle{splncs04}
\bibliography{main}
\end{document}